\documentclass{elsarticle}
\usepackage{times}
\usepackage{graphicx}
\usepackage{latexsym}
\usepackage{booktabs}
\usepackage{makecell}
\usepackage{multirow}
\usepackage{arydshln}
\usepackage{amsmath}

\usepackage{lineno,hyperref}
\modulolinenumbers[5]

\usepackage{microtype}
\usepackage{enumitem}
\usepackage{tabularx}
\usepackage{tikz}
\usepackage{pgfplots}
\usepackage{subfigure}
\usepackage{graphicx}
\usepackage{MnSymbol}
\usepackage{footmisc}
\usepackage[colorinlistoftodos]{todonotes}

\usepackage{xurl}

\journal{Journal of \LaTeX\ Templates}

\begin{document}

\begin{frontmatter}

\title{Search-Engine-augmented Dialogue Response Generation \\ with Cheaply Supervised Query Production}

\author{Ante Wang$^{a,*}$, Linfeng Song$^{b,*}$, Qi Liu$^c$, Haitao Mi$^b$, Longyue Wang$^b$, Zhaopeng Tu$^b$, Jinsong Su$^{a,\dagger}$, Dong Yu$^b$}
\address{$^a$Xiamen University, $^b$Tencent AI Lab, $^c$University of Oxford}
\address{wangante@stu.xmu.edu.cn, lfsong@tencent.com, jssu@xmu.edu.cn}

\begin{abstract}
Knowledge-aided dialogue response generation aims at augmenting chatbots with relevant external knowledge in the hope of generating more informative responses.
The majority of previous work assumes that the relevant knowledge is given as input or retrieved from a static pool of knowledge. However, this assumption violates the real-world situation, where knowledge is continually updated and a chatbot has to \emph{dynamically} retrieve useful knowledge.
We propose a dialogue model that can access the vast and dynamic information from any search engine for response generation. 
As the \emph{core} module, a query producer is used to generate queries from a dialogue context to interact with a search engine.
We design a training algorithm using \emph{cheap noisy supervision} for the query producer, where the signals are obtained by comparing retrieved articles with the next dialogue response.
As the result, the query producer is adjusted \emph{without} any human annotation of gold queries, making it easily transferable to other domains and search engines.
Experiments show that our query producer can achieve R@$1$ and R@$5$ rates of 62.4\% and 74.8\% for retrieving gold knowledge, and the overall model generates better responses over strong knowledge-aided baselines using BART \cite{lewis2020bart} and other typical systems.

\renewcommand{\thefootnote}{\fnsymbol{footnote}} 
\footnotetext[1]{Both authors contributed equally to this work.} 
\footnotetext[2]{Corresponding author.} 

\end{abstract}


\end{frontmatter}


\section{Introduction}

The task of knowledge-aided dialogue response generation aims to find useful knowledge for an on-going conversation to help a chatbot generate more relevant and engaging responses.
This is an important direction for dialogue response generation due to three advantages: (1) it allows a dialogue model to access a large pool of knowledge beyond local conversational contexts; (2) it enables a dialogue model to capture the dynamic nature of the world \cite{komeili2021internet}, where knowledge sources are frequently updated; (3) it may enhance the interpretability of dialogue models by examining retrieved knowledge and allows fine-grained interventions by replacing certain pieces of knowledge \cite{adiwardana2020towards,zhang2020dialogpt,roller2021recipes}.

\begin{figure}[t]
    \centering
    \includegraphics[width=0.7\textwidth]{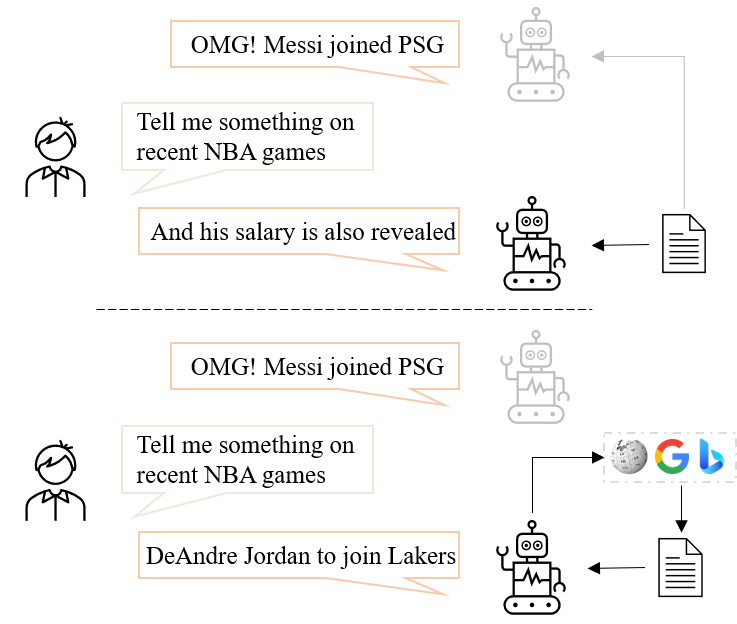}
    \caption{Previous knowledge-aided dialogue response generation models (up), where related articles are given as input, versus our model (down), which can dynamically fetch knowledge using a standard search engine.}
    \label{fig:intro}
\end{figure}

Initial efforts \cite{ghazvininejad2018knowledge,liu2018knowledge,wu2019proactive,zhou2020kdconv,tian2020response,chen2020bridging,Kim2020Sequential} on knowledge-aided response generation assume that relevant knowledge (e.g., news or movie reviews) is given as input and design dialogue systems that can effectively utilize the provided knowledge.
However, as shown in Fig. \ref{fig:intro}, this static setting violates the dynamic nature of real-world scenarios. This gives rise to approaches that can retrieve and select information from a knowledge source for response generation \cite{zhao2020knowledge,dinan2018wizard,lee2019latent}. These projects assume searching from a static pool of articles (e.g., a Wikipedia dump). The queries and articles are represented as sparse vectors of $n$-grams \cite{dinan2018wizard} or even dense contextualized vectors \cite{lee2019latent} for retrieval. However, these approaches with a static pool of knowledge still fall short of taking the dynamic nature of knowledge into account.

\begin{figure*}[t]
    \centering
    \includegraphics[width=0.99\textwidth]{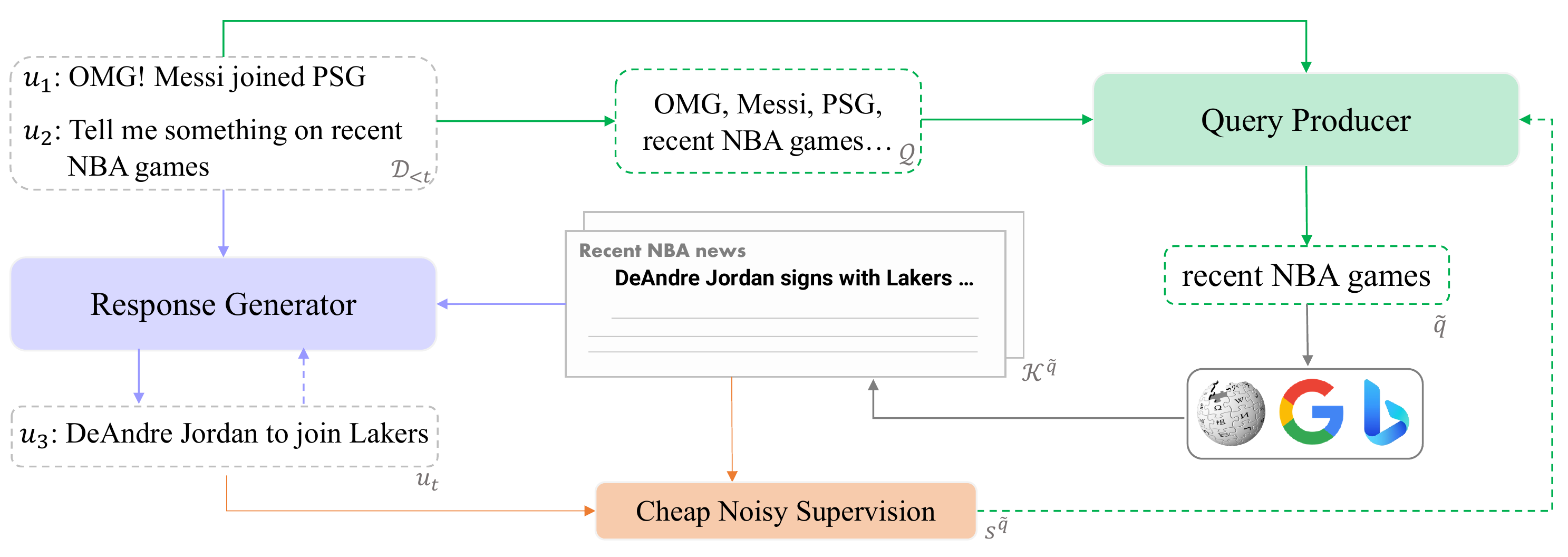}
    \caption{The training process using the example in Fig. \ref{fig:intro}, where solid lines 
    and dashed lines 
    indicate forward and backward pass. \textbf{First} (\textcolor{green}{$\rightarrow$}), input utterances $\mathcal{D}_{<t}$ and (optional) query candidates $\mathcal{Q}$ are fed into the \colorbox{green!20}{query producer} to get search query $\tilde{q}$, and then (\textcolor{gray}{$\rightarrow$}) relevant articles $\mathcal{K}^{\tilde{q}}$ are retrieved from a search engine with $\tilde{q}$. \textbf{Next} (\textcolor{blue}{$\rightarrow$}), the \colorbox{blue!20}{response generator} constructs the next dialogue turn $u_t$ given both $\mathcal{D}_{<t}$ and $\mathcal{K}^{\tilde{q}}$. \textbf{Finally} (\textcolor{orange}{$\rightarrow$}), supervision signals are calculated based on $\mathcal{K}^{\tilde{q}}$ and $u_t$ to update (\textcolor{green}{$\dashedrightarrow$}) the query producer. The response generator is updated (\textcolor{blue}{$\dashedrightarrow$}) based on a cross-entropy loss over $u_t$.}
    \label{fig:pipeline}
\end{figure*}

In this paper, we propose a dialogue model that can access the vast and dynamic knowledge from any search engine for response generation. We choose to work with search engines based on two reasons. First, search engines (e.g., Google) store continually updating knowledge, which well captures the dynamic nature of our world. Second, we get rid of the difficulties of building our own search engines with $n$-grams and dense contextualized vectors, since the ranking algorithms of well-established search engines are highly optimized. Fig. \ref{fig:pipeline} shows the framework of our model, consisting of a query producer and a response generator. The query producer generates queries from a dialogue context. Then, we send the queries to a search engine to obtain relevant articles. The response generator takes both the retrieved articles and the dialogue context to generate a response.

As a key component in our model, the query producer determines the quality of fetched knowledge, which further affects response generation. 
However, annotating gold queries are costly, because annotators usually need to examine multiple candidate queries by looking into their fetched articles.
To obtain automatic training signals for our query producer, we design a function based on existing cheap noisy supervision for scoring queries.
It simply compares the retrieved articles of a query with the corresponding gold response to estimate the quality of the query.
The scoring function does not require extra annotations, such as gold queries, making our model easily transferable to other domains and search engines.

We use Wizard of Wikipedia (WoW, \cite{dinan2018wizard}), a popular benchmark on knowledge-aided response generation, for evaluating our model, taking the publicly free search engine from Wikipedia to retrieve knowledge instead of using the static knowledge provided by WoW. 
Experiments show that our query producer can achieve a R@$1$ (R@$5$) rate of 62.4\% (74.8\%) for retrieving the correct knowledge on the \emph{unseen} test set of WoW. Besides, our model generates better replies than a strong BART \cite{lewis2020bart} model and knowledge-aided baselines with heuristic algorithms for query acquisition. These results indicate the feasibility of using a search engine as the knowledge source for response generation.
\footnote{Our source code is available at \url{https://github.com/DeepLearnXMU/SEA-DialogGen}.}

\section{Model}

Formally, given a dialogue context of prior $t-1$ turns $\mathcal{D}_{<t}=\{u_1, u_2,...,u_{t-1}\}$, our model first predicts a query $\tilde{q}$ (optionally from a set of query candidates $\mathcal{Q}=\{q^1,q^2,...,q^{|\mathcal{Q}|}\}$ selected by a heuristic algorithm), before sending it to a search engine for retrieving a list of articles $\mathcal{K}^{\tilde{q}}=\{k^{\tilde{q}}_1,k^{\tilde{q}}_2,...,k^{\tilde{q}}_{|\mathcal{K}^{\tilde{q}}|}\}$. With the retrieved knowledge $\mathcal{K}^{\tilde{q}}$ and dialogue context $\mathcal{D}_{<t}$, a response $u_t$ is generated. 

Fig. \ref{fig:pipeline} visualizes the workflow of our model. In the rest of this section, we introduce the two key components, the query producer (\S \ref{sec:QP}) and the response generator (\S \ref{sec:RG}). 


\subsection{Query Production}
\label{sec:QP}
We explore two popular directions based on either extraction (\S \ref{sec:EP}) or generation (\S \ref{sec:GP}) to build our query producer. We further prune the query search space to minimize the number of possible queries and speed up training (\S \ref{sec:SSP}). We use cheap noisy supervisions to train the query producers with Maximum Likelihood Estimation (MLE) based pre-training and reinforcement learning fine-tuning (\S \ref{sec:f_func}).

\subsubsection{QP-Ext: Extraction-based Query Producer}
\label{sec:EP}
Our extraction-based query producer aims to extract text spans from the dialogue context $\mathcal{D}_{<t}$ as queries. We use a pre-trained language model (\texttt{PLM}) as its backbone and add a linear layer with the softmax activation (\texttt{MLP-Softmax}) as the output layer to predict the probability distribution $\boldsymbol{\mathrm{P}}$ over all query candidates $\mathcal{Q}=[q^1,\dots,q^{|\mathcal{Q}|}]$:
\begin{equation}
\begin{aligned}
    \boldsymbol{\mathrm{P}}&={\rm MLP\text{-}Softmax}([\boldsymbol{\mathrm{H}}^{q^1},...,\boldsymbol{\mathrm{H}}^{q^{|\mathcal{Q}|}}]), \\
    \boldsymbol{\mathrm{H}}^{q^i}&={\rm MeanPooling}(\boldsymbol{\mathrm{H}}_{beg_i:end_i}), \\
    \boldsymbol{\mathrm{H}}&={\rm PLM}(\mathcal{D}_{<t}) \text{,}
\end{aligned}
\label{eq:ext_score}
\end{equation}
where $\boldsymbol{\mathrm{H}}$ represents the contextualized embeddings produced by \texttt{PLM}, and $beg_i$ and $end_i$ are the begin and end indices for the $i$-th candidate span in $\mathcal{D}_{<t}$.
Each candidate query $q^i$ is a continuous span in a turn of $\mathcal{D}_{<t}$. We use ${\rm MeanPooling}$ over the contextualized embeddings of its tokens from $beg_i$ to $end_i$ to get its representation $\boldsymbol{\mathrm{H}}^{q^i}$.

\subsubsection{QP-Gen: Generation-based Query Producer}
\label{sec:GP}
Different from the extraction-based model, this generation-based model adopts a seq2seq architecture to construct search queries from scratch.
It can produce queries that are not contained in $\mathcal{D}_{<t}$ at the cost of a larger search space.
We adopt a pre-trained encoder-decoder model (denoted as \texttt{PGM}) to generate queries in an auto-regressive manner, and beam search is adopted during decoding to produce multiple queries at the same time \cite{meng-etal-2017-deep}. 
The score $s_i$ for a query $q^i$ is the sum of the log probabilities for its tokens over the whole vocabulary:
\begin{equation}
\begin{aligned}
    s_i &= \frac{\sum_{j=1}^{|q^i|} \log {\rm MLP\text{-}Softmax}(\boldsymbol{\mathrm{H}}^{q^i}_j)}{\sqrt{|q^{i}|}} , \\
    \boldsymbol{\mathrm{H}}^{q^i}_j &= {\rm PGM}(\mathcal{D}_{<t}, q^{i}_{<j}) \text{,}
\end{aligned}
\label{eq:gen_score}
\end{equation}
where $\boldsymbol{\mathrm{H}}^{q^i}_j$ is the decoder state of the $j$-th step for query $q_i$, and $\sqrt{|q_i|}$ is the length-based normalization item to ease the preference of short candidates.

\subsubsection{Pruning Query Search Space}
\label{sec:SSP}
Querying a search engine can be time consuming for training a query producer, as the training process can take hundreds of thousands of steps, and each query can take more than 0.1 seconds.
A natural solution for this issue is to create an offline cache of articles for all possible queries before the actual training.
However, both extraction-based and generation-based models take a large search space of candidate queries.
Given a dialogue of $m$ turns with $n$ words for each turn, there are $\mathcal{O}(m\cdot n^2)$ possible queries for the extraction-based model, while the number is exponential to average query length for the generation-based model.

We study different methods to prune the search space for query production, so that an offline cache can be efficiently established, while the coverage of the pruned space is still large enough. In particular, we explore the two main directions in the task of keyword acquisition \cite{siddiqi2015keyword}.
\begin{itemize}[leftmargin=*]
\item \emph{Dictionary-based}: Typical methods in this direction \cite{ferragina2010tagme} consider the overlap between each dialogue context and a predefined taxonomy as the search space, where the taxonomy is constructed from a large knowledge source (e.g. Wikipedia).
\item \emph{Metric-based}: Approaches in this direction \cite{rose2010automatic,CAMPOS2020257} extract keywords from a dialogue context based on metric scores (e.g., TF-IDF) without using any vocabulary, and then they merge adjacent keywords into larger spans by heuristic rules.
\end{itemize}

\subsubsection{Training with Cheap Noisy Supervision}
\label{sec:f_func}

We leverage a \emph{cheap noisy supervision} signal to train our query producers, which makes it easier to transfer to other domains and search engines compared with using human annotations \cite{komeili2021internet}.
The whole training process contains \emph{pre-training with cross-entropy loss} and \emph{reinforcement learning fine-tuning}. The reinforcement learning fine-tuning directly uses the supervision signals as reward, while the pre-training uses the signals as gold labels.

\subparagraph{Cheap noisy supervision for query scoring}
\label{sec:RD}
We design a function $f$ that leverages the corresponding gold response $u$ as cheap noisy supervision to assign a score $s_q$ for each query $q$ to indicate its quality.
In particular, the function $f$ compares the corresponding top articles $\mathcal{K}^q=\{k_1^q,k_2^q,\dots\}$ retrieved by $q$ with the gold response $u$ for calculating score $s^q$: 
\begin{equation} \label{eq:f_func}
    s^q = f(\mathcal{K}^q, u) \text{.}
\end{equation}
We consider this as a type of \emph{cheap} supervision because the function $f$ \emph{does not} require extra annotations (e.g., the annotations of gold queries).
We study different approaches and choose the popular BM25 metric \cite{robertson1994some} to implement $f$.
More specifically, it first calculates the score for each article by $s_i^q={\rm BM25}(k_i^q,u)$, before determining the overall score $s^q$ as the maximum among them: $s^q=\max(\{s_1^q,s_2^q,\dots\})$.

We introduce two pre-processing methods for improving upon the vanilla BM25.
The first method adopts coreference resolution, which finds the actual entity referred by a pronoun. We then expand response $u$ by concatenating it with the entity mentions referred by its pronouns. This is important as coreference frequently exists in human conversations. The second method drops function words from both articles $\mathcal{K}$ and response $u$ before passing them to the noisy supervision function $f$. This makes $f$ focus more on content words.

\subparagraph{Pre-training with noisy labels}
At this stage, we take the query with the highest score $s^q$ by function $f$ (Eq. \ref{eq:f_func}) from query candidates $\mathcal{Q}$ as pseudo ground-truth to train both extraction-based and generation-based producers with the standard cross-entropy loss: 
\begin{align}
    \mathcal{L}_{ext.}^{pt}&=-\log P(\bar{q}|\mathcal{D}_{<t},\theta_{ext.}), \\
    \mathcal{L}_{gen.}^{pt}&=-\sum_{i=1}^{|\bar{q}|} \log P(\bar{q}_i|\mathcal{D}_{<t},\bar{q}_{<i},\theta_{gen.}) \text{,} \label{eq:ce_gen}
\end{align}
where $\bar{q}$ denotes the pseudo ground-truth, $\mathcal{L}_{ext.}^{pt}$ and $\mathcal{L}_{gen.}^{pt}$ are loss terms for extraction-based and generation-based models respectively, and $\theta_{ext.}$ and $\theta_{gen.}$ are the parameters for the models.

\subparagraph{Fine-tuning using reinforcement learning}
At the fine-tuning stage, we adopt a REINFORCE algorithm \cite{williams1992simple} with the cheap noisy supervision $f$ as the reward. We subtract a baseline value, which is set to the reward of the candidate query with the highest model score (calculated by Eq. \ref{eq:ext_score} or \ref{eq:gen_score}) from $f$ to reduce variance. As BM25 scores are not bounded, we further normalize them to reduce training variance. For each dialog turn with multiple query candidates, we rescale the  reward $r_i$ for the $i$-th candidate as $\frac{r_i-min}{max-min}-0.5$ with the minimum ($min$) and maximum ($max$) values within the candidates.
The losses for both producers at fine-tuning stage are defined as:
\begin{align}
    \mathcal{L}^{ft} &= - \Delta(r_s,r_b) \log p_s \text{,}
\end{align}
where $p_s$ is the probability of a candidate query sampled from the model output distribution, $r_s$ and $r_b$ are the rescaled rewards for the sampled and the baseline candidates, respectively. 

\subsection{Response Generation}
\label{sec:RG}
After retrieving relevant articles, the next step of our model is to generate a proper response using the articles and the dialogue context. We implement response generators, Rank-Gen and Merge-Gen, based on two representative research directions.
Both models use different strategies to leverage the retrieved articles, and thus we can better study the robustness of our query producer. 



\subsubsection{Rank-Gen}
\label{sec:RaGe}

Rank-Gen takes an explicit ranker to choose one piece from a set of articles \cite{ijcai2019-0706,NEURIPS2020_6b493230,zhao2020knowledge}.
There are several benefits of this direction, such as improving the explainability and the ability of handling large knowledge set.
The ranker first selects a piece of knowledge $\tilde{k}$ from candidates $\mathcal{K}$, then the seq2seq-based generator predicts the response given the dialogue context $\mathcal{D}_{<t}$ and selected knowledge $\tilde{k}$:
\begin{equation}
\begin{aligned}
    \tilde{k} &= {\rm argmax}_{k \in \mathcal{K}}{\rm Ranker}(\mathcal{D}_{<t}, k), \\
    u_t &= {\rm Generator}(\mathcal{D}_{<t}, \tilde{k}).
\end{aligned}
\end{equation}

We adopt reinforcement learning to jointly train the ranker and generator, where the ranker is guided by the signal from the generator via policy gradient, and the generator is trained by cross-entropy loss taking sampled knowledge $\tilde{k}_s$ from the ranker:
\begin{align}
    \mathcal{L}_{RG} &= \mathcal{L}_{rank} + \mathcal{L}_{gen},\\
    \mathcal{L}_{rank} &= -(\mathcal{L}^{\tilde{k}_b}_{gen} - \mathcal{L}^{\tilde{k}_s}_{gen}) \log P(\tilde{k}_s|\mathcal{D}_{<t}, \mathcal{K}) , \\
    \mathcal{L}_{gen} &= - \sum_{i=1}^{|u_t|} \log P(u_{t,i}|u_{t,<i}, \mathcal{D}_{<t}, \tilde{k}_s) \text{,}  
\end{align}
where $\tilde{k}_b$ is the baseline knowledge to reduce variance, and $\mathcal{L}_{gen}^{x} (x\in \{\tilde{k}_b,\tilde{k}_s\})$ is the generation loss taking the corresponding knowledge as extra input.

Before joint training, we also introduce a warm up stage following \cite{zhao2020knowledge}, where the ranker is trained with cross-entropy loss on the pseudo ground-truth knowledge $\bar{k}$ that has the highest BM25 score among knowledge candidates, and the generator is also trained with cross-entropy loss taking $\bar{k}$ as the additional input:
\begin{align}
    \bar{k} &= {\rm argmax}_{k \in \mathcal{K}}{\rm BM25}(\mathcal{D}_{<t}, \mathcal{K}), \\
    \mathcal{L}_{rank}^{pt} &= - \log P(\bar{k}|\mathcal{D}_{<t}, \mathcal{K}), \\
    \mathcal{L}_{gen}^{pt} &= - \sum_{i=1}^{|u_t|} \log P(u_{t,i}|u_{t,<i}, \mathcal{D}_{<t}, \bar{k}) \text{.}  
\end{align}




\subsubsection{Merge-Gen}
\label{sec:Merge-Gen}

Merge-Gen is our implementation of the FiD model \cite{izacard2021leveraging,shuster2021retrieval}, which
follows another popular direction by consuming all input knowledge.
Particularly, each knowledge piece $k_i$ in knowledge pool $\mathcal{K}$ is first paired with the dialogue context $\mathcal{D}_{<t}$. Then, these pairs $\{\mathcal{D}_{<t}, k_i\}_{k_i \in \mathcal{K}}$ are encoded into hidden states independently before being concatenated as inputs to the decoder for response generation:
\begin{equation}
\begin{aligned}
    u_t &= {\rm Decoder}([\boldsymbol{\mathrm{H}}_1;\boldsymbol{\mathrm{H}}_2;...;\boldsymbol{\mathrm{H}}_{|\mathcal{K}|}]), \\
    \boldsymbol{\mathrm{H}}_i &= {\rm Encoder}(\mathcal{D}_{<t}, k_i) \text{.}  
\end{aligned}
\end{equation}
Comparing with Rank-Gen, Merge-Gen does not suffer from the risk of selecting wrong knowledge by a ranker.
However, it lacks explainability and is costly for its decoder to process fused hidden states of all input knowledge.
The training signal is based on the standard cross-entropy loss over gold response $u_t$:
\begin{equation}
    \mathcal{L}_{MG} = - \sum_{i\in[1..|u_t|]} \log P(u_{t,i}|u_{t,<i}, \mathcal{D}_{<t}, \mathcal{K}) \text{.}  
\end{equation}

\section{Experiment}
We study the effectiveness of our model, especially the usefulness of knowledge retrieval using search queries for response generation.

\subsection{Dataset}
We choose the Wizard-of-Wikipedia (WoW, \cite{dinan2018wizard}) dataset for evaluation. 
The dataset is split into 18,430/967/968 dialogues for train/dev/test, respectively. 
For each dialogue, it includes the relevant knowledge (e.g., the title of the ground-truth article) annotated by human.
Therefore, we can use WoW to \emph{additionally} measure the performance of query production by comparing the titles of a retrieved article and the ground-truth article.
We use its \emph{unseen} test set for evaluation. We remove the first turn of each dialogue, because the first turn reveals the title of the Wikipedia article for discussion, which will expose the main topic of the dialogue.

\subparagraph{Search engine} 

We choose Wikipedia Search\footnote{\url{https://en.wikipedia.org/wiki/Special:Search}}, a free vertical search engine that returns the latest content of Wikipedia\footnote{Though Wikipedia seems to be static, it is in fact dynamically updated. According to \url{https://en.wikipedia.org/wiki/Wikipedia:Statistics}, it develops at a rate of around 2 edits every second, and the English Wikipedia alone gets 585 new articles per day.} given a user query.
We retain the \textbf{top 5} retrieved Wikipedia articles for each query for response generation,
extracting the summary of each article (the first paragraph of a Wikipedia article) as external knowledge.

\subsection{Setting}
We choose hyperparameters by following previous work or the results of development experiments.

\subparagraph{Query production}
The official ELECTRA-base \cite{Clark2020ELECTRA:}\footnote{\url{https://huggingface.co/google/electra-base-discriminator}} 
and BART-base \cite{lewis2020bart}\footnote{\url{https://huggingface.co/facebook/bart-base}} 
models 
are taken as the backbones for extraction-based and generation-based query producers, respectively.
We use AdamW \cite{loshchilov2018decoupled} with learning rate $10^{-5}$ and batch size 64 to optimize our models.
The extraction-based producer is pre-trained for 1 epoch, while the generation-based producer is pre-trained for 5 epochs. To prune the search space of query production, we adopt two keyword acquisition tools, TagMe (dictionary-based) 
and YAKE! (metric-based). We use recall, denoted as R@$x$ ($x\in \{1,3,5\}$), which compares the top $x$ retrieved candidates with ground-truth knowledge to evaluate the performance of query producers.

\subparagraph{Response generation}
Both Rank-Gen and Merge-Gen use a BART-base model for response generation. All models are trained using AdamW with learning rate 1e-5 and batch size 64. The warm-up stage for ranker in Rank-Gen takes 2 epoch. We perform early stopping based on the perplexity (PPL) on the development set. Following previous work, We adopt PPL and Unigram F1 to evaluate response generation.

\subsection{Development Experiments}
We explore the design choices for query space pruning (\S \ref{sec:SSP}) and the scoring function $f$ (Eq. \ref{eq:f_func}), as they determine the quality of query production, which in turn affects response generation.

\label{sec:dev1}
\begin{table}[t]
    \centering
    \begin{tabular}{llccc}
    \toprule
    Pruning & Query Scoring
    & R@1 & R@3 & R@5 \\
    \midrule
    \multirow{6}{*}{\makecell[c]{TagMe}}
    & Random & 12.55 & 31.27 & 44.19 \\
    & TF-IDF & 39.30 & 61.28 & 67.26 \\
    & BM25$(q,u)$ & 36.09 & 58.73 & 65.89 \\
    & BM25 & 53.36 & 65.25 & 69.46 \\
    & BM25$_{++}$ & \textbf{60.59} & \textbf{69.81} & \textbf{72.49} \\
    \midrule
    \multirow{6}{*}{\makecell[c]{YAKE!}}
    & Random & 14.21 & 33.96 & 46.00 \\
    & TF-IDF & 36.92 & 58.63 & 64.78 \\
    & BM25$(q,u)$ & 28.01 & 52.94 & 62.59 \\
    & BM25 & 50.70 & 65.32 & 69.91 \\
    & BM25$_{++}$ & 57.97 & 69.15 & 72.03 \\
    \bottomrule
    \end{tabular}
    \caption{Development results of various search-space pruning methods and query scoring algorithms.}
    \label{tab:dev_algo}
\end{table}

\subparagraph{Different choices of space pruning and query scoring algorithms}
Table \ref{tab:dev_algo} shows the development results of several popular query scoring algorithms with \emph{TagMe}  and \emph{YAKE!} for search space pruning. Among the scoring algorithms:
\begin{itemize}[leftmargin=*]
\setlength{\itemsep}{0pt}
\item \emph{Random}: It randomly picks a query from the candidate pool.
\item \emph{TF-IDF}: It averages the TF-IDF scores of all words within a candidate query as its score. This algorithm \emph{only considers the query information}.
\item \emph{BM25($q$,$u$)}: It measures the similarity between $q$ and $u$ using BM25 without considering the actual retrieved knowledge by $q$.
\item \emph{BM25}: It is our proposed scoring function $f$ (Eq. \ref{eq:f_func}) with standard BM25.
\item \emph{BM25$_{++}$}: It is also based on $f$ using BM25 but equipped with pre-processing methods: coreference resolution and function words dropping.
\end{itemize}

\begin{figure}[t]
\centering
\pgfplotsset{every axis/.append style={
semithick},
}
\begin{tikzpicture}[scale=0.8]
\begin{axis}[
    legend style={
    at={(0.98,0.38)},
    cells={anchor=west}},
    xtick={1,2,3,4,5,6,7,8,9},
    xticklabels={1,2,3,4,5,6,7,8,$\ge$9},
    xlabel=Turns,
    ylabel=R@$x$(\%)]
\addplot[sharp plot,color=red, mark=*] coordinates {
    (1,44.50)(2,57.71)(3,59.32)(4,60.06)(5,60.34)(6,60.54)(7,60.41)(8,60.54)(9,60.59)
    };
\addlegendentry{R@$1$}
\addplot[dashed,color=blue, mark=triangle*] coordinates {
    (1,48.60)(2,65.53)(3,67.32)(4,68.95)(5,69.38)(6,69.63)(7,69.71)(8,69.84)(9,69.81)
    };
\addlegendentry{R@$3$}
\addplot[dash dot,color=green, mark=square*] coordinates {
    (1,48.91)(2,67.14)(3,69.46)(4,71.37)(5,71.70)(6,72.28)(7,72.31)(8,72.44)(9,72.49)
    };
\addlegendentry{R@$5$}
\addplot[dash dot dot,color=cyan, mark=diamond*] coordinates {
    (1,48.96)(2,67.88)(3,70.73)(4,73.33)(5,74.17)(6,74.99)(7,75.21)(8,75.44)(9,75.47)
    };
\addlegendentry{R@M}
\end{axis}
\end{tikzpicture}
\caption{Development results of BM25$_{++}$ and the ceiling performances (R@M) given keyword candidates from the last $k$ turns.}
\label{fig:dev_knum}
\end{figure}
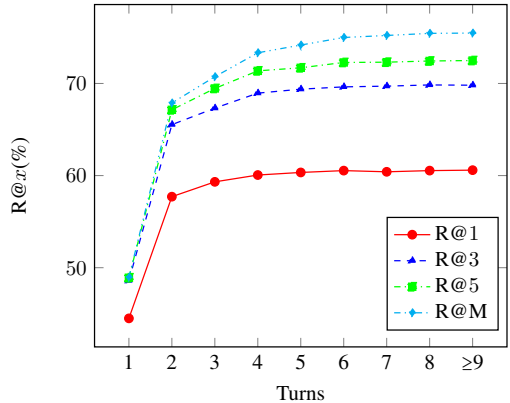

Regarding search-space pruning, the average candidate number and the ceiling performance (R@M in Fig. \ref{fig:dev_knum}) using TagMe are 17.45 and 75.47\%, respectively, while the corresponding numbers are 21.64 and 75.04\% for YAKE!.
\textbf{First}, the upper bound does not reach 100\% because: (1) the pruning method fails to keep some good search queries; (2) some dialogue turns (4.7\%) do not require any external knowledge; (3) speakers change the topics in some turns, which requires queries that are not contained in the dialogue context. Overall, we get a decent number of around 75\%. \textbf{Second}, most ranking algorithms using TagMe outperform their corresponding ones using YAKE!. Besides, TagMe reaches higher upper bound (75.47\% vs 75.04\%) with less candidates (17.45 vs 21.64) than YAKE!.
Based on the results, we choose TagMe for query space pruning in further experiments.

Regarding query scoring, BM25$_{++}$ outperforms all other algorithms, demonstrating the effectiveness of coreference resolution and function words dropping. BM25 is the second best method, which shows that the retrieved articles provide more information beyond the query and the response. We choose BM25$_{++}$ for future experiments.


\subparagraph{The number of dialogue turns for obtaining candidate queries}
With the pruning method and query scoring algorithm determined, the next step is to choose the number ($k$) of turns for obtaining candidate queries.
Intuitively, considering more turns will increase the ceiling performance on knowledge retrieval with extra noise on the query scoring algorithm.
As shown in Fig. \ref{fig:dev_knum}, the performance of BM25$_{++}$ consistently improves with the increase of $k$, indicating that the benefit of considering longer dialogue context for candidate queries exceeds the cost (extra noise).
Therefore, we choose to consider all turns for the remaining experiments.


\begin{table*}[t]
\setlength\tabcolsep{4pt}
    \centering
    \begin{tabular}{lc|ccc|cc|cc}
    \toprule
    \multirow{2}{*}{\makecell[c]{Query/KN \\ Production}} 
    & \multirow{2}{*}{\makecell[c]{Avg. Num. \\ Querying}}
    & \multicolumn{3}{c|}{Query Ranking} 
    & \multicolumn{2}{c|}{Rank-Gen} 
    & \multicolumn{2}{c}{Merge-Gen} \\
    &
    & R@1 & R@3 & R@5
    & PPL$\downarrow$ & Uni. F1 
    & PPL$\downarrow$ & Uni. F1 \\
    \midrule
    None & -- & -- & -- & -- 
    & 25.26 & 16.53 & 25.13 & 16.64 \\
    \hline
    Last 2 turns & 8.29
    & -- & -- & -- 
    & 22.77 & 17.43 
    & 20.04 & 17.55 \\
    Last 4 turns & 13.38 
    & -- & -- & -- 
    & 22.86 & 17.38 
    & 19.89 & 17.72 \\
    All history & 17.45
    & -- & -- & -- 
    & 23.03 & 17.32 
    & \textbf{19.79} & 17.71 \\
    \hline
    Concat & \multirow{1}{*}{\textbf{1}}
    & \;\;4.15  & \;\;5.45 & \;\;5.88  
    & 24.79 & 16.76 
    & 24.57 & 16.51 \\
    TF-IDF & \multirow{1}{*}{\textbf{1}}
    & 43.41 & 61.63 & 66.65 
    & 22.86 & 17.28 
    & 21.53 & 17.64 \\
    QP-Ext & \textbf{1}
    & \textbf{62.41} & \textbf{72.91} & \textbf{74.87}
    & \textbf{21.60} & \textbf{17.81}
    & 20.20 & \textbf{18.15} \\
    QP-Gen & \textbf{1}
    & 56.77 & 66.08 & 68.22
    & 21.65 & 17.51
    & 20.69 & 17.95 \\
    \midrule
    \midrule
    GENRE \cite{decao2020multilingual} 
    & -- & -- & -- & -- 
    & 22.59 & 17.60 
    & 20.24 & 18.15 \\
    Gold KN & -- & -- & -- & -- 
    & 18.83 & 18.63 
    & 18.99 & 18.42 \\
    \bottomrule
    \end{tabular}
    \caption{Main results of query production and response generation on WoW unseen test set, where ``PPL$\downarrow$'' and ``Uni. F1'' indicates perplexity and unigram F1, respectively. \emph{QP-Ext} is significantly better than \emph{TF-IDF} at $p<0.01$ across all aspects, and \emph{QP-Gen} is better than \emph{TF-IDF} at $p<0.05$.} 
    \label{tab:main_exp}
\end{table*}

\subsection{Main Results}

Table \ref{tab:main_exp} shows the main testing results including the performance on search query production and response generation. 
We compared our models with typical baselines with different query acquisition techniques:
(1) no external knowledge is used (first group); (2) fetching knowledge with all search queries in the last $k$ turns that are not filtered by query pruning\footnote{They are based on the heuristic that people tend to keep talking the topics just mentioned in the last few turns.} (second group); (3) using search queries produced by different techniques (third group); (4) several upper bounds to pinpoint the current bottleneck of our model (last group).
We design the baselines in the second group to better highlight the merit of our model.
This is because the queries in later turns are more likely to retrieve gold knowledge than those in earlier turns, as people tend to keep discussing the topics they just mention.
For the third group, \emph{Concat} concatenates each dialogue context as the query, 
\emph{TF-IDF} uses the TF-IDF score to pick the query, 
and both \emph{QP-Ext} (\S \ref{sec:EP}) and \emph{QP-Gen} (\S \ref{sec:GP}) correspond to our query production methods.

We can draw the following conclusions: 
\textbf{First}, leveraging external knowledge is always helpful for dialogue response generation (comparing the first line with others).
\textbf{Second}, Merge-Gen based models tend to perform better than the corresponding Rank-Gen based ones, because it avoids the error propagation from the ranker and uses multiple pieces of knowledge.
Besides, for the baselines using multiple queries (the second group), Rank-Gen and Merge-Gen show opposite trends when the number of turns for obtaining queries increases with Merge-Gen being consistently better.
Both results confirm the advantage of Merge-Gen over Rank-Gen in two aspects: Merge-gen can utilize multiple pieces of knowledge at the same time; and it prevents the error propagation from using an explicit ranker.
\textbf{Third}, using more queries for knowledge fetching can generally improve the overall performance, but the time of knowledge gathering (querying a search engine and retrieving pages) also grows linearly with the query number. 
For instance, the querying time can be more than 2 seconds when using 10 queries.
\textbf{Lastly}, our models using either of the proposed query producers perform better than all baselines for most situations, indicating that our query producer trained with cheap noisy supervision signals can retrieve useful contents for response generation.
The \emph{Concat} baseline fails to get any articles for many cases, because its generated queries are very long.
The baselines (the second group) using multiple queries show slightly better perplexity values than our models when combined with Merge-Gen.
But, their knowledge fetching process is at least 8-time slower than ours, \emph{causing delay} in response time and being \emph{computational costly}.
Besides, our models still manage to get better Uni. F1 scores with fewer times of search-engine querying.
We also observe a positive correlation between query ranking performance and response generation performance, which again validates the necessity of studying query production.

We also compare several \emph{upper-bound} systems (the last group) to pinpoint the current bottleneck of our model.
The first system, \emph{GENRE} \cite{decao2020multilingual}, is pretrained on the WoW dataset to generate the title of the corresponding Wikipedia article given a dialogue context.
Thus, it utilizes the human annotated labels during training.
Besides, it adopts constrained decoding for inference so that all generation outputs are valid Wikipedia titles.
For \emph{GENRE}, we use the proposed ``constrained beam search'' to get 5 distinct titles, which are then mapped to 5 different passages (same number as other systems) as the knowledge for response generation.
It yields a recall number of 67.55\% on gold articles, while the R@1 of \emph{QP-Ext} is 62.41\%.\footnote{This is roughly comparable as we use the top 5 articles for each query.}
\emph{GENRE} can also be considered as a purely retrieval-based model that directly get articles from a static Wikipedia dump. 
Although, \emph{GENRE} is trained with the human annotation on knowledge selection, \emph{QP-Ext} is comparable on both knowledge retrieval and the final response generation, indicating the potential of search-engine-based approaches and our training method with cheap noisy supervision.

The other upper-bound system, \emph{Gold KN}, takes the summary (the first paragraph) of the gold article for both training and testing.
We observe significant performance gaps on response generation
when either Rank-Gen or Merge-Gen is used.
This indicates the potential for further improving the accuracy on query production, which in turns boosts the recall of knowledge retrieval.

\subsection{Analysis}

\begin{table}[t]
    \centering
    \begin{tabular}{lccc}
    \toprule
    System & R@1 & R@3 & R@5 \\
    \midrule
    \makecell[l]{QP-Ext} & 62.41 & 72.91 & 74.87 \\
    \makecell[l]{~~~~w/o pre-train} 
    & 61.97 & 71.84 & 73.77 \\
    \makecell[l]{~~~~w/o fine-tune} 
    & 61.36 & 73.08 & 74.94 \\
    \makecell[l]{~~~~w/o prune search space} 
    & 60.65 & 67.68 & 69.97 \\
    \hline
    \makecell[l]{QP-Gen}
    & 56.77 & 66.08 & 68.22 \\
    \makecell[l]{~~~~w/o pre-train} 
    & 38.14 & 54.91 & 59.83 \\
    \makecell[l]{~~~~w/o fine-tune} 
    & 51.91 & 65.82 & 69.75 \\
    \makecell[l]{~~~~w/ prune search space} 
    & 60.67 & 71.55 & 73.52 \\
    \bottomrule
    \end{tabular}
    \caption{Ablation study on both extraction-based (QP-Ext) and generation-based (QP-Gen) query producers.}
    \label{tab:ablation}
\end{table}

\subparagraph{Ablation study}
Table \ref{tab:ablation} shows the ablation study on our query producers. We can draw the following conclusions.
\textbf{First}, both pre-training with cross-entropy loss and reinforcement learning fine-tuning are helpful for query producers. 
For extraction-based approach, pre-training (w/o fine-tune) mainly helps the performance on R@3 and R@5, while fine-tuning (w/o pre-train) mostly helps the performance on R@1.
In general, fine-tuning provides more robust performances than pre-training, as it can better handle the noisy supervision.
For generation-based method, both training stages are very crucial, probably due to its large search space.
In this case, pre-training-alone (w/o fine-tune) outperforms the fine-tuning-alone counterpart (w/o pre-train).
This is because RL-based fine-tuning from scratch is slow to converge \cite{paulus2018a,wang2018reinforced}.
\textbf{Second}, adding search space pruning brings in significant performance gains on both extraction-based and generation-based query producers, proving the importance of limiting the search space to high-quality candidate queries.
Notice that for a generation-based query producer, it will degenerate to a ranking model when using search space pruning, because its search space is limited within provided query candidates. Though effective, it loses the ability to generate queries not contained in the dialogue context, which has great potential for further improvement. We leave it as future work.

\begin{figure}
\centering
    \begin{tikzpicture}[scale=0.74]
    \begin{axis}[
        legend style={
        cells={anchor=west}},
        xlabel=Turns,
        ylabel=R@1(\%)]
    \addplot[dash dot dot,color=blue, mark=triangle*] coordinates {
        (2,65.75)(3,42.76)(4,48.04)(5,40.78)(6,42.57)
        (7,38.37)(8,39.45)(9,28.97)(10,29.80)
        };
    \addlegendentry{TF-IDF}
    \addplot[sharp plot,color=red, mark=*] coordinates {
        (2,77.29)(3,70.39)(4,70.89)(5,64.69)(6,62.10)
        (7,55.04)(8,53.90)(9,44.89)(10,43.91)
        };
    \addlegendentry{Ext. based}
    \addplot[dash dot,color=green, mark=square*] coordinates {
        (2,70.89)(3,62.28)(4,66.40)(5,55.92)(6,57.81)
        (7,49.56)(8,50.39)(9,42.24)(10,38.82)
        };
    \addlegendentry{Gen. based}
    \end{axis}
    \end{tikzpicture}
    \caption{Performance of different query producers at different dialogue turns.}
    \label{fig:conv_flow}
\end{figure}
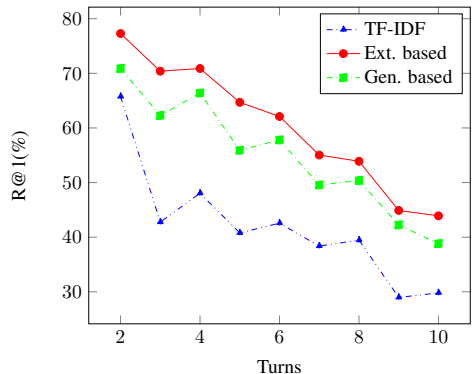

\textbf{Performances at different turns}~
We further compare the R@$1$ of query producers at various turns.
Generally, the last several turns yield more query candidates than the first ones, causing larger search spaces.
As shown in Fig. \ref{fig:conv_flow}, the performance of all producers drops rapidly when a dialogue continues.
But, ours still achieve R@1 rates of around 45\% and 40\% with only noisy supervision, which are 15\% and 10\% higher than TF-IDF.

\subsection{Human Evaluation}

\begin{table}
    \centering
    \setlength{\tabcolsep}{4pt}
    \begin{tabular}{lc|c|cc}
         \toprule
         \multirow{2}{*}{\makecell[c]{Query Prod.}} 
         & Query & Article & \multicolumn{2}{c}{Response} \\
         & Sound. & KN Cov.
         & Natural. & Know. \\
         \midrule
         None & -- & -- & 2.39  & 1.89 \\
         TF-IDF & 2.37 & 2.37 & 2.41 & 2.14 \\
         QP-Ext & \textbf{2.79} & \textbf{2.76} & \textbf{2.65} & 2.39 \\
         QP-Gen & 2.65 & 2.59 & 2.58  & \textbf{2.44} \\
         \midrule
         \midrule
         GENRE \cite{decao2020multilingual} & 2.78 & 2.85 & 2.57 & 2.67 \\
         \bottomrule
    \end{tabular}
    \caption{Human evaluation results, where \emph{Sound.}, \emph{KN Cov.}, \emph{Natural.}, \emph{Know.} indicates soundness, knowledge coverage, naturalness and knowledgeable respectively. Both \emph{QP-Ext} and \emph{QP-Gen} are significantly better over TF-IDF at $p<0.01$ across all aspects.}
    \label{tab:human_eval}
\end{table}

We conduct human evaluation on 100 test samples, and we choose Merge-Gen as the response generator, because it shows better performance than Rank-Gen on automatic metrics.
The models are rated regarding query production, the quality of retrieved article, and response generation.
For query production, we measure \textbf{Soundness}, which means whether the query is sound by itself.\footnote{Sometimes, a sound query may not retrieve good knowledge due to search-engine mistake.}
For the quality of retrieved article, we evaluate its \textbf{Coverage}, meaning how relevant it is to the current dialogue context.
For response generation, we follow previous work to measure \textbf{Naturalness}, indicating how fluent and relevant a response is, and \textbf{Knowledgeable}, representing how much knowledge is used in a response.
We ask 3 annotators capable of fluent English communication to score each aspect with 3-point schema\footnote{We attach detailed guidelines in Appendix.}, and we average their scores as the final score of the aspect.
The inner-annotator agreement (Fleiss' $\kappa$) is 0.5461, indicating a moderate level of agreements.

As shown in Table \ref{tab:human_eval}, both the \emph{TF-IDF} baseline and our models (\emph{QP-Ext} and \emph{QP-gen}) improve the \emph{None} baseline regarding the ``knowledgeable'' aspect of response generation, which indicates that using retrieved knowledge is helpful.
Comparatively, our \emph{QP-Ext} model significantly improves \emph{TF-IDF} ($+$0.42 for ``query soundness'' and $+$0.39 for ``knowledge coverage'' over 3-point schema) regarding the quality of obtained queries and articles.
This advantage also transfers to response generation ($+$0.24 for ``naturalness'' and $+$0.25 for ``knowledgable'' over 3-point schema), which also indicates the positive correlation between query production and the final response generation.
The improvements (especially for \emph{TF-IDF}) regarding ``naturalness'' are relatively low, because \emph{None} can already generate fluent replies aided by large-scale pre-training on text generation.
Note that general replies such as ``\emph{Sorry, I don't know}'' are considered natural in certain context like ``\emph{Do you know Mike Tyson?}''.

We also compare with the upper-bound system, \emph{GENRE} \cite{decao2020multilingual}, for a more comprehensive understanding of our models.
Generally, our \emph{QP-Ext} shows comparable ``soundness'' but lower ``knowledge coverage'' score, which also leads to inferior ``knowledgeable'' score.
This is because \emph{GENRE} can provide knowledge across various topics by directly producing the passage titles, 
but our \emph{QP-Ext} retrieves knowledge with much similar topic using only 1 query.
This indicates that using more queries and smarter knowledge pre-filtering can further benefit the query production.
We also notice that \emph{GENRE} give a lower ``naturalness'' score than our \emph{QP-Ext}.
It may be because the response generator enjoys the high quality knowledge from \emph{GENRE} but instead ignores the dialogue context information.
This shows that the current model still struggles to balance the ``naturalness'' and ``knowledgeable''.

\begin{table*}[t!] \scriptsize
    \centering
    \begin{tabularx}{0.98\textwidth}{rX}
        \textbf{\#1} \\
    \midrule
        Context & \textbf{A:} One of my favorite bands despite usual protests is Nickelback.  How about you or who is one of your favorites? \\
        & \textbf{B:} I like Nickelback too. They are my favorite Canadian rock band and I've liked them since 1995 they started. \\
        & \textbf{A:} Speaking of Canadian rock bands, my absolute all time favorite band is Rush. I have every album by them! \\ 
        Ref response & Yes Rush! They formed in 1968 and took their first United States tour in 1974. \\
    \midrule
        \textbf{None} & Rush is one of my favorite bands too. They are from Hanna, Alberta. \\
    \midrule
        \textbf{TF-IDF} \\
        \hdashline
        Seach query & Nickelback \\
        Retrieved passages & Nickelback; Rockstar (Nickelback song); Nickelback discography; Chad Kroeger; Dark Horse (Nickelback album) \\
        Knowledge fragment & ...Nickelback is one of the most commercially successful Canadian rock bands, having sold more than 50 million albums worldwide... \\
        Rank-Gen & Rush is one of my favorite bands too. They have sold more than 50 million albums worldwide. \\
        Merge-Gen & Rush is one of my all time favorite bands. They have sold more than 50 million albums worldwide. \\
    \midrule
        \textbf{QP-Ext} \\
        \hdashline
        Seach query & Rush \\
        Retrieved passages & Rush (Rush album); Into the Rush; Rush (band) \\
        Knowledge fragment & ...Rush was a Canadian rock band formed in Toronto in 1968, consisting of Geddy Lee (bass, vocals, keyboards, composer), Alex Lifeson (guitars, composer), and Neil Peart (drums, percussion, lyricist)... \\
        Rank-Gen & Rush is one of my favorite bands too. They were formed in 1968 in Toronto. \\
        Merge-Gen & Rush was formed in 1968 by Geddy Lee, Alex Lifeson, and Neil Peart. \\
    \end{tabularx}
    \caption{The first example (1/2) regarding query production and response generation from the WoW unseen test set.}
    \label{tab:example1}
\end{table*}

\begin{table*}[t!] \scriptsize
    \centering
    \begin{tabularx}{0.98\textwidth}{rX}
        \textbf{\#2} \\
        \midrule
        Context & \textbf{A:} Do you like Elvis Presley? he was a famous american singer \\
        & \textbf{B:} He is pretty good, I am from Nashville, So i've even been to graceland. \\
        Ref response & Presley was born in Tupelo, Mississippi, and then to memphis \\
    \midrule
        \textbf{No KN} & He was born and raised in Nashville, TN \\
    \midrule
        \textbf{TF-IDF} \\
        \hdashline
        Seach query & graceland \\
        Retrieved passages & Graceland; Graceland (album); Graceland (TV series); Graceland University \\
        Knowledge fragment & ...Graceland is located at 3764 Elvis Presley Boulevard in the vast White haven community, about nine miles (14 kilometers) from Downtown and less than four miles (6.4 km) north of the Mississippi border.It was opened to the public as a museum on June 7, 1982. The site was listed in... \\
        Rank-Gen & Yes, it was opened to the public as a museum in 1982 \\
        Merge-Gen & I've been to Graceland a few times. It was opened to the public as a museum in 1982 \\
    \midrule
        \textbf{QP-Ext} \\
        \hdashline
        Seach query & Elvis Presley \\
        Retrieved passages & Elvis Presley; Elvis Presley (album); Priscilla Presley; Elvis Presley albums discography; Lisa Marie Presley \\
        Knowledge fragment & ...Presley was born in Tupelo, Mississippi, and relocated to Memphis, Tennessee with his family when he was 13... \\
        Rank-Gen & He was born in Tupelo, Mississippi and moved to Memphis when he was 13 \\
        Merge-Gen & He was born in nashville and moved to Memphis when he was 13 \\
    \end{tabularx}
    \caption{The second example (2/2) regarding query production and response generation from the WoW unseen test set.}
    \label{tab:example2}
\end{table*}

\subsection{Case Study}

We further demonstrate several typical examples from our human study to help visualize the benefits of our query producer for response generation, as shown in Table \ref{tab:example1} and \ref{tab:example2}. We compare our models with a vanilla BART and models equipped with a TF-IDF query producer because all these models are trained without annotated queries.
The baseline (NO KN) suffers from hallucination problem where the generated responses in these cases are conflicted with the facts. For example, Elvis Presley is actually born in Tupelo, Mississippi instead of Nashville, TN.
Another baseline (TF-IDF) generate responses that tally with the facts. However, they fail to produce correct queries as it is difficult for TF-IDF to consider the rich contextual information and recognize the most related topic.
Our models using knowledge from QP-Ext correctly produce the search queries and generate the most satisfying responses thanks to our cheaply supervised training framework. 
But we notice that the retrieved passages may not cover the fact used in gold reference (e.g., ``Rush took their first United States tour in 1974'' in the first example) and the model may still suffer from hallucination given necessary facts (e.g., ``Presley was born in nashville'' in the second example). These problems need further exploring in future study.

\section{Related Work}

\subparagraph{Knowledge-aided dialogue response generation}
How to properly incorporate external knowledge has become an important topic in dialogue response generation.
Regarding the type of adopted knowledge, previous work has explored using passages \cite{dinan2018wizard,zhou2018dataset,gopalakrishnan2019topical}, knowledge graphs \cite{moon2019opendialkg,zhou2020kdconv}, commonsense knowledge \cite{zhou2018commonsense,zhang2020grounded,wu2020diverse}, persona \cite{li2016persona,zhang2018personalizing,madotto2019personalizing} as the external knowledge, and recent efforts \cite{moghe2018towards,wu2021more} even propose integrating multiple sources of knowledge.
These efforts mainly focus on the ``knowledge-centric'' scenario where each dialogue mainly discusses the corresponding knowledge (e.g., a short passage), and thus simply using the knowledge can enjoy a high coverage on the dialogue.
However, a dialogue model may need to dynamically fetch relevant knowledge in practical scenarios (e.g., open-domain chitchat), as the discussed topic may change through time.
Later work \cite{ijcai2019-0706,zhao2020knowledge,shuster2021retrieval,chi2021neural,saha2021proto} constructs retrieval-based generative models, where they adopt a retriever (e.g. DPR \cite{karpukhin2020dense}) to obtain relevant knowledge from a static knowledge pool.
Though they can potentially access more knowledge, the knowledge pool remains unchanged.

To tackle the above issue, 
search-engine-aided dialogue response generation is recently proposed so that the vast knowledge form internet can be leveraged.
This is an important yet under explored research direction with the key being query production, which is to generate search queries to interact with a search engine.
There is one related preprint draft \cite{komeili2021internet} in parallel, which explores using Bing\footnote{\url{https://www.bing.com/}} as the knowledge source for dialogue response generation.
We both share a similar motivation of using a search engine as the knowledge source, but we are different on how to train the query producer, a \emph{key} module for interacting with the search engine.
\cite{komeili2021internet} manually annotate 48K queries to train their query generator.
Thus, the supervision signals are expensive to obtain and may not be transferable to other domains and search engines.
On the other hand, our model is search-engine agnostic, as we design an algorithm that obtains annotation-free and effective signals for training the query producer.
In addition, \cite{komeili2021internet} ignores the fact that Bing (same as the other search engines) is frequently updated, thus the same query may not result in the same retrieved articles after a long while.
Comparatively, it is much less costly to update our model for handling web content update.

We also notice several very recent preprint drafts \cite{lazaridou2022internet,menick2022teaching} that propose to leverage internet for language model pretraining or question answering.
Our work share a similar motivation on leveraging internet to solve a practical task.
However, all these efforts simply annotate queries to train their query producer, while we study leveraging cheap noisy supervision.

\subparagraph{Keyword production}
As a longstanding task, keyword production was initially proposed to automatically create keywords for articles. Classic techniques (e.g., TF-IDF and TextRank) have been widely used over decades. 
Initial neural keyword producers \cite{zhang2016keyphrase,luan2017scientific} are extraction-based that extract keywords from inputs.
Recently, generation-based methods \cite{meng-etal-2017-deep,chen2018keyphrase,Chen_Gao_Zhang_King_Lyu_2019,meng-etal-2021-empirical,xie2022wr} using a seq2seq model are gaining popularity. 
We produce keywords as queries to a search engine and study both extraction-based and generation-based methods on our task in conversational domain.

\subparagraph{Cheaply Supervised Training}
Training a model with supervision from human annotations is the commonly-used approach to build the model for a task. 
However, it is costly to collect enough annotations to train a strong model and it is still challenging when serving cross-domain settings.
Thus, many researchers have explored self-supervised pretraining losses as cheap supervision for exploiting common knowledge for various downstream tasks \cite{lewis2020bart,Clark2020ELECTRA:,devlin2019bert}.
Some work \cite{su2021enhanced,jiang2021exploring} also adopt self-training or reinforcement learning free of human annotations to enhance model abilities.
In this work, we are the first to design cheap noisy supervision for conversational query producers on knowledge-aided dialogue response generation task.

\section{Conclusion}
We have introduced a model that leverages a general search engine for knowledge-aided response generation.
To effectively interact with the search engine, it adopts a query producer to generate search queries.
We design cheap noisy supervision signals to train our query producer, so that no extra human annotation is needed, making our model easily transferable to other search engines and domains.
Experimental results under both automatic metrics and human judges show the superiority of our model over a pre-trained BART model and other baselines.

\section*{Acknowledgments}
This work was supported by National Natural Science Foundation of China (No. 62276219), 
Natural Science Foundation of Fujian Province of China (No. 2020J06001),
and Youth Innovation Fund of Xiamen (No. 3502Z20206059).

\bibliography{mybibfile}

\appendix

\clearpage

\clearpage


\section{Annotation Guidelines}

All aspects are based on a 3-point scheme: 3 means flawless; 2 means containing minor flaw; 1 means having major flaw but with values; 0 means being completely wrong.

\paragraph{Query Soundness}
It considers if the selected topic is active (the one being discussed).
\begin{itemize}[leftmargin=*]
\setlength{\itemsep}{0pt}
    \item The score can be 3 if the active topic is selected, otherwise the score can be 2, 1 or 0 depends on how close the selected topic is to the active one. 
    \item If the active topic (e.g., ``plants vs zombie'') is emerged from a parent topic (e.g., ``zombie''), choosing the parent topic (``zombie'') can yield a score of 2.
\end{itemize}

\paragraph{Article Knowledge Coverage} It measures how relevant (and useful) are the retrieved articles regardless of the query (sometimes a bad query can yield good articles).
\begin{itemize}[leftmargin=*]
\setlength{\itemsep}{0pt}
    \item If the article talks about something (e.g., guitars) close to the dialogue topic (e.g., a guitarist), then the score can be 2.
    \item If the article (e.g., about guitars) is slightly relevant to the dialogue topic (e.g., a musician who is not a friend of the guitarist), the score can be 1.
    \item The score can be 0 if no article is retrieved (sometimes this is due to bad queries).
\end{itemize}

\paragraph{Naturalness} 
How sound a reply is to the dialogue context.
A sound reply should be consistent both in purpose and in topic to the context.
But it does not reflect the knowledge aspect (e.g., if it violates the provided knowledge).
\begin{itemize}[leftmargin=*]
\setlength{\itemsep}{0pt}
    \item If there is a question like ``Do you like ...?'', a sound reply should contain something like ``Yes...'', ``No, I don't...'' or ``I do...''
\end{itemize}

\paragraph{Knowledgeable} 
A knowledgeable reply should contain new stuff, so examples like ``Oh, that's cool!'' is not knowledgeable. In this situation, scores can range from 0 to 1, where 1 can be chosen if the reply actually does not require knowledge.

Besides, knowledgeable replies should not violate factoid statements in both dialogue context and in retrieved knowledge. For instance, if the context mentions ``the band sold 500 million albums worldwide'', it is not knowledgeable if the reply says ``the band sold 400 million albums worldwide''.
\begin{itemize}[leftmargin=*]
\setlength{\itemsep}{0pt}
\item For a context asking for knowledge, if a reply (e.g., ``Oh, that's cool!'') does not contain any knowledge, the score can be 0.
\item For replies that violate existing factoid statements, the score can be 1.
\item For replies that cannot be determined true or false given dialogue context and retrieved knowledge, the score can be 2.
\item For replies that can be found true given dialogue context and retrieved knowledge, the score can be 3.
\end{itemize}




\end{document}